\documentclass{article}
\usepackage{spconf,amsmath,epsfig}

\providecommand{\doi}[1]{doi: {\footnotesize \href{http://dx.doi.org/#1}{\path{#1}}}}

\usepackage{tabularx}
\usepackage{multirow}
\usepackage{pgfplots}
\usepackage{caption}
\usepackage{subcaption}

\usepackage[pdftex=true,breaklinks=true,hidelinks=true,colorlinks=true,citecolor=blue]{hyperref}

\usepackage[square,numbers]{natbib}

\def\x{{\mathbf x}}

\title{Autonomous Payload Thermal Control}
\name{Alejandro D. Mousist}
\address{Thales Alenia Space in Spain}
\begin{document}
\ninept
\maketitle
%
\begin{abstract}
In small satellites there is less room for heat control equipment, scientific instruments, and electronic components. Furthermore, the near proximity of electronic components makes power dissipation difficult, with the risk of not being able to control the temperature appropriately, reducing component lifetime and mission performance.
To address this challenge, taking advantage of the advent of increasing intelligence on board satellites, an autonomous thermal control tool that uses deep reinforcement learning is proposed for learning the thermal control policy onboard.
The tool was evaluated in a real space edge processing computer that will be used in a demonstration payload hosted in the International Space Station (ISS).
The experiment results show that the proposed framework is able to learn to control the payload processing power to maintain the temperature under operational ranges, complementing traditional thermal control systems.
\end{abstract}




\section{Introduction}
In satellites, thermal control is a critical task that consists of maintaining the temperature of all its components within an acceptable range. Sensors, batteries, optics, joints, and bearings are some of the most affected components when the temperature goes below or above the operational range.

In small satellites, there is less room for heat control equipment, scientific instruments, and electronic components. In addition, the proximity of the electronic components makes it challenging to dissipate power. High temperatures do activate many degradation pathways that reduce component lifetime \cite{electronics8121423}.

Generally speaking, thermal control is achieved primarily by controlling heat fluxes, and the techniques for doing so are classified as active or passive thermal control systems. SmallSats typically favor passive systems over active ones because of their lower cost, volume, weight, and risk.
For components with more stringent temperature limitations or greater heat loads, active thermal management approaches, however, have proven to be more successful in maintaining closer temperature control. Unfortunately, due to power, mass, and volume requirements, the bulk of them are challenging to integrate into small satellites \cite{nasa}.

With the advent of onboard intelligence in SmallSats, it is increasingly feasible to assist tasks like navigation, calibration, or fault detection in place using tools like artificial intelligence which enable fast reaction and adaptation in an adverse environment like space. Because the heat created by the spacecraft is one of the primary contributors to the spacecraft's temperature, active management of it is a potential challenge to be tackled on board.

The spacecraft may learn a thermal control strategy by interacting with the environment. It might, among other things, turn on or off non-critical systems, alter clock speed, parallelize or serialize processing, halt or postpone processing, and so on to control power consumption and, as a result, adjust the emitted thermal radiation adaptively to different external situations (e.g., eclipses, direct sunlight, albedo variations).

It has been proposed in prior publications to employ model-based approaches like fuzzy \cite{wang2021research, dong2012fuzzy, 5365072} or Proportional Integral Derivative (PID) \cite{noauthor_2019-lb} controllers to accomplish adaptive heat regulation. 
These approaches aim to model thermal dynamics with simplified mathematical models, but they have some drawbacks. On the one hand, they struggle to accurately simulate the complexities of thermal dynamics and the various influencing factors \cite{gao2019energy}. On the other hand, fuzzy control systems require expert knowledge to define control rules, whereas PID requires manual adjustment of key parameters, which can be time-consuming and hinder adaptive online adjustments. In \cite{xiong2020intelligent}, an actor-critic deep reinforcement learning (DRL) algorithm is used to solve a live tuning problem for a PID controlling the heater voltage of a space telescope.

In this work, an on-board system called \emph{APaTheCSys} (\textbf{A}utonomous \textbf{Pa}yload \textbf{The}rmal \textbf{C}ontrol \textbf{Sys}tem) that uses DRL for helping in the thermal control of the payload is proposed, letting the agents to learn the thermal behavior of the payload on each scenario in a model-free manner. The system is evaluated in real hardware that has been included in the conceptual space edge computing mission called IMAGIN-e\footnote{IMAGIN-e is a demostration space edge computing payload that will be hosted in the International Space Station, carried out in a collaboration agreement by Thales Alenia Space and Microsoft}.

\section{Materials and Methods}
\subsection{Problem Definition}
The system to be controlled consists of a processing node that contains a System on Chip (SoC) based on ARM architecture with 16 cores that can be managed independently. The system temperature is monitored using nine thermistors, seven of which are scattered throughout the SoC and the other two are located on the top and bottom edges of the main board.

Given a maximum temperature limit, the agent learns to manage the hardware resources available for processing in order to maximize performance while maintaining the overall system temperature below the upper limit.

The agent interacts with the environment using a standard Application Programming Interface (API) created with the Gymnasium \cite{gymnasium} library.

A containerized version of the \emph{stress-ng} tool was used for generating a sustained load and controlled from the Gymnasium scenario setup.

While interacting with the environment, the agent explores sub-optimal policies that, in certain situations, may allow the temperature to scale to values that are not suitable for some components. In this experimentation the load was controlled and was removed in such cases but in a real mission, a rigid thermal control should be used in conjunction (e.g. powering-off the processing node, i.e. the SoC, when its temperature is out of range).

\begin{figure}[ht]
    \centering
    \includegraphics[width=8cm]{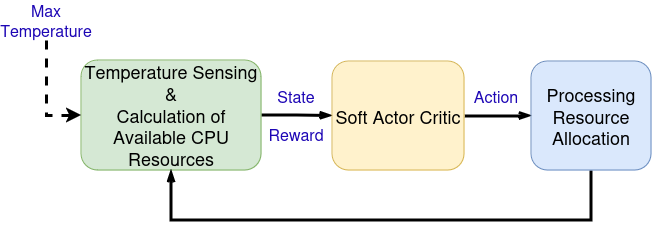}
    \caption{Architecture scheme for the on-board payload power control }
    \label{fig:general}
\end{figure}

The temperature control system proposed in this work is schematized in Fig. \ref{fig:general}.

\subsection{Deep Reinforcement Learning Modeling}
\subsubsection{Actions}
The agent is able to control the processing power by modifying both the power status and clock speed for each available Central Processing Unit (CPU).

The power status of the CPUs can be managed using the CPU hotplug feature which allows to switch cores on or off dynamically using the sysfs interface.

The CPU clock speed can be managed via the CPUFreq subsystem which allows the operating system to scale the CPU frequency up or down to save power or improve performance.

The actions to be carried out by the agent are intended to establish the states listed in Table \ref{table:cpu_status} on each CPU.

\begin{table}[h]\renewcommand{\arraystretch}{1}
\begin{center}
\begin{tabular}{|c|c|}
\hline\hline
Power Status & Clock Speed\\
\hline\hline
On & Low (1MHz)\\
On & High (up to 2MHz)\\
Off & ------\\
\hline\hline
\end{tabular}
\caption{Available options for setting on each CPU in terms of power and clock speed}
\label{table:cpu_status}
\end{center}
\end{table}

Although CPUs can be uniquely identified by their number and their state modified independently, in this study the agent controls the state of the CPUs without distinction between them. This model reduces the action space and therefore increases the learning performance as suggested by \cite{booth2019ppo}.

Accordingly, the agent is allowed to choose the following quantities from a reduced and continuous action space:
\begin{itemize}
\item \textbf{Powered-on CPUs}: Percentage of the total amount of CPUs present in the board that are powered-on.
\item \textbf{High frequency CPUs}: Percentage of the powered-on CPUs that are in high clock speed
\end{itemize}

\subsubsection{State}
\begin{figure}[ht]
\centering
\resizebox{0.4\textwidth}{0.3\textwidth}{%
\begin{tikzpicture}[declare function={func(\y) = 46+(2+16/8)*ln(2*\y);}]
\begin{axis}[
    xmin = 0.5,
    xmax = 1.5,
    ymin = 46,
    ymax = 52,
    xlabel={Time},
    ylabel={Temperature},
    ytick={48.465,48.772588722, 50},
    xtick={0.93,1},
    yticklabels={$T(t_{-1})$,$T(t)$, $T_{limit}$},
    xticklabels={$t_{-1}$, $t$},
    height=.3\textheight,
    axis lines = middle
    ]
    \addplot[domain = 0.5:1,smooth,thick,magenta, line width=1.5pt] {46+(2+16/8)*ln(2*\x)};

    \draw[dotted, red] (0,247) -- (43,247);
    \node at (43,247)[circle,fill,inner sep=1.2pt]{};
    \draw[dotted, red] (43,0) -- (43,247);

    \draw[dotted, red] (0,277) -- (50,277);
    \node at (50,277)[circle,fill,inner sep=1.2pt]{};
    \draw[dotted, red] (50,0) -- (50,277);

    \addplot[domain = 0.7:1.15,smooth,semithick,dashed, black] {4.0201343 * \x + 44.752454422} node [pos=1, above right] {$m$};;

    \node[draw,align=left] at (80,130) {$m= \frac{T(t)-T({t-1})}{t-t_{-1}}$};
    \draw[dotted, thick, black] (0,400) -- (82,400);

    \draw[dotted, thick, blue] (0,277) -- (82,277);
    \draw [decorate,decoration={brace,amplitude=5pt,mirror,raise=4pt},yshift=0pt] (80,277) -- (80,400) node [black,midway,xshift=0.8cm] {$\Delta T$};
\end{axis}
\end{tikzpicture}
}
\caption{Representation of the \textbf{\emph{margin to limit}} and \textbf{\emph{slope}} quantities which are sent to the agent as part of the state data.} \label{fig:state}
\end{figure}
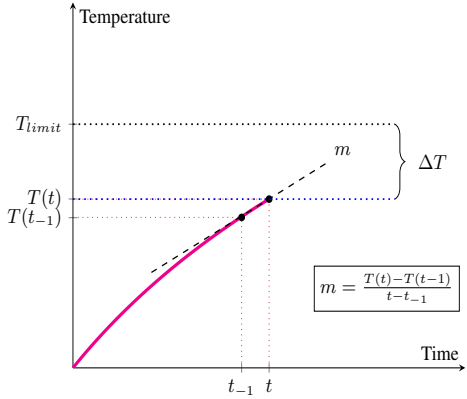

The agent receives a normalized continuous state that includes the following quantities:
\begin{itemize}
\item \textbf{CPU state ($P$)}: This parameter is calculated in equation \ref{eq:parameter} and reflects the current processing power considering the power status (on/off) and frequency level assigned (low/high) on each CPU.

\begin{equation} \label{eq:parameter}
    P=\frac{\sum_{cpu=1}^{n} P_{cpu}}{2*n}
\end{equation}

\[
P_{cpu}= \left\{ \begin{array}{lcl}
          0 &  if  &  \textrm{cpu=off} \\
          1 &  if  &  \textrm{cpu=on and frequency=low} \\
          2 &  if  &  \textrm{cpu=on and frequency=high} 
          \end{array}
\right.
\]

In the equation, \emph{n} represents the number of available CPUs (16 in the SoC used).

\item \textbf{Margin to limit ($\Delta T=T_{limit} - T$)}: Represents the available margin before reaching the temperature limit ($T_{limit}$) considering the current temperature of the system ($T$) (Represented in Fig. \ref{fig:state}).
\item \textbf{Slope ($m$)}: Expresses the rate of change of temperature. Instead of the derivative of an unknown temperature function, this quantity is replaced with the slope of a secant line that is calculated using real temperature measures in a brief period (Represented in Fig. \ref{fig:state}).
\end{itemize}

\subsubsection{Rewards}
The goal is to operate as many CPUs at the highest frequency possible while maintaining the required temperature range.
Positive rewards are accumulated by the agent while the temperature is below the temperature threshold, and a negative reward is given when this threshold is achieved to allow the agent to learn while interacting with the environment.

On each step, if the threshold temperature has not been reached, the reward is calculated with Equation \ref{eq:reward}.

\begin{equation} \label{eq:reward}
    R=\frac{cpus_{high}*C_{high} + cpus_{low} * C_{low} + cpus_{off}* C_{off}}{n}
\end{equation}

As in previous equations, \emph{n} represents the number of available CPUs. $C_{high}$, $C_{low}$ and $C_{off}$ are constants used to weigh independently each term and to express how each state (power status and clock speed of the CPUs) conditions the reward.

\subsubsection{Algorithm}
Many interactions with the environment are necessary for the DRL algorithms to learn to behave correctly. Due to this low efficiency in the use of experience, it is still challenging to use reinforcement learning in real life problems.
Off-policy algorithms collect samples in a replay buffer that is used on every policy update making them ideal in cases where gaining experience has a high cost because they can efficiently learn from past events and decisions.

In this study, a model-free, off-policy algorithm called Soft Actor-Critic \cite{haarnoja2018soft} (SAC) was used to train the agent. SAC is an actor-critic algorithm, and as such follows a policy-based and value-based approach, that concurrently learns a policy and two Q-functions. 
The Stable Baselines3 (SB3) \cite{stable-baselines3} implementation of SAC was used in the experimentation. SB3 is a set of implementations of reinforcement learning algorithms in Pytorch that uses the Gymnasium library as an abstraction for the environments.

\subsubsection{Pre-training: Temperature Modeling} \label{modelling}
To mitigate the impact of initial DRL algorithm iterations on the real environment, a simulated environment has been employed. This simulation aims to replicate the thermal behavior of the board. While the model only captures certain aspects of the actual environment, it enables the agent to explore the connection between available temperature margin, temperature curve slope, and allocated processing power in a harmless context.

The agent can learn and subsequently adapt its behavior in relation to the real scenario. This is sometimes referred to as domain adaptation.

A family of temperature functions over time conditioned not only by time but also by a variable determined by the state of the CPU was proposed as a naive modeling of the temperature curve followed by the system.

Equation \ref{eq:temperature_estimator} represents the family of equations used in temperature evolution estimation.

\begin{equation} 
    Y(x)=T_{init} + (2 + \frac{P}{8}) \ln(2x)
\label{eq:temperature_estimator}
\end{equation}

In equation \ref{eq:temperature_estimator} the parameter \emph{$P$}, also included in the state data equation (\ref{eq:parameter}),  determines a concrete equation in relation to the CPU state.

\section{Results and Discussion}

\subsection{Experimental results}
In the experiment, the limit temperature was established at 55 degrees Celsius since this threshold was close to the operating limit of various subsystems.

The policies pre-trained on a software simulated environment were compared with others learned from scratch on the hardware environment. After initial episodes of domain adaptation, the pre-trained policies were able to maintain the system operating below the temperature threshold for more than 3 hours while the policies trained from scratch on the hardware environment need multiple days of training for obtaining equivalent results. Due to this, the rest of the experimentation was done using pre-trained policies.

\begin{figure}
     \centering
     \begin{subfigure}[b]{0.5\textwidth}
         \centering
         \includegraphics[width=\linewidth]{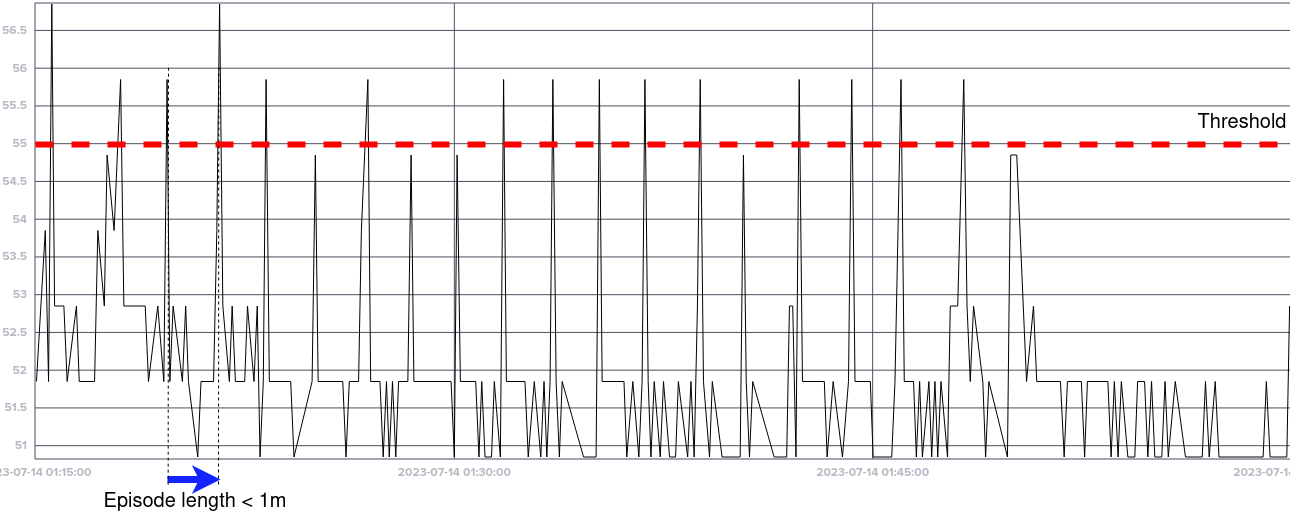}
         \caption{Duration of the episodes at the beginning of the training.}
         \label{fig:beginning_training}
     \end{subfigure}
     \hfill
     \begin{subfigure}[b]{0.5\textwidth}
         \centering
         \includegraphics[width=\linewidth]{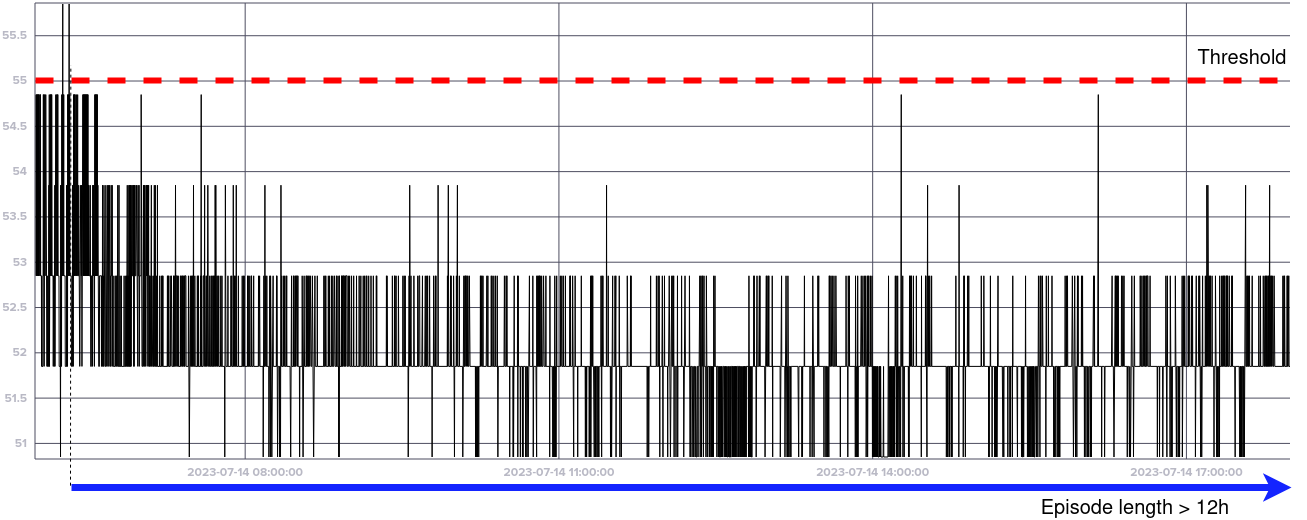}
         \caption{Duration of the episodes after training the 33rd episode.}
         \label{fig:5h_training}
     \end{subfigure}
     \hfill
        \caption{Comparison of the episode duration at the beginning and after initial training phase.}
        \label{fig:comparison}
\end{figure}

Fig. \ref{fig:beginning_training} shows a sample of the initial episode cycles in temperature graphic of the board. In this episode, the maximum temperature threshold was reached in less than 5 iterations (25 seconds).

After training for 30 episodes, the agent was able to maintain the temperature below the limit temperature for more than 12h, as shown in Fig. \ref{fig:5h_training}, illustrating the improvement in the agent skills.

\begin{figure}
\centering
\begin{tikzpicture}
\begin{axis}[%
xmin=0,
xmax=57,
xtick distance=10,
ymin=0,
ymax=5500,
ytick distance=500,
xlabel=Episode,
ylabel=Length (Iterations),
height=.28\textheight
]
\addplot [
color=red,
solid,
mark=o,
patch type=cubic spline,
mark options={scale=1.0, color=blue},
every node near coord/.style={text=black, anchor=south west},
forget plot
]
table[row sep=crcr]{
1	14\\
2	10\\
3	13\\
4	12\\
5	8\\
6	7\\
7	24\\
8	6\\
9	7\\
10	5\\
11	489\\
12	50\\
13	56\\
14	22\\
15	1642\\
16	2674\\
17	75\\
18	59\\
19	49\\
20	699\\
21	1190\\
22	3232\\
23	47\\
24	89\\
25	68\\
26	4988\\
27	830\\
28	49\\
29	3826\\
30	479\\
31	33\\
32	19\\
33	41\\
34	5001\\
35	3721\\
36	2919\\
37	3308\\
38	1622\\
39	58\\
40	2562\\
41	358\\
42	5001\\
43	657\\
44	215\\
45	3896\\
46	455\\
47	667\\
48	725\\
49	5001\\
50	1581\\
51	1792\\
52	2004\\
53	3306\\
54  5001\\
55  5001\\
56  749\\
57  385\\
};
\end{axis}
\end{tikzpicture}%
    \caption{Evolution of the episode's lengths during training. Episodes max length is limited to 5000 iterations.}
    \label{fig:hw_training}
\end{figure}
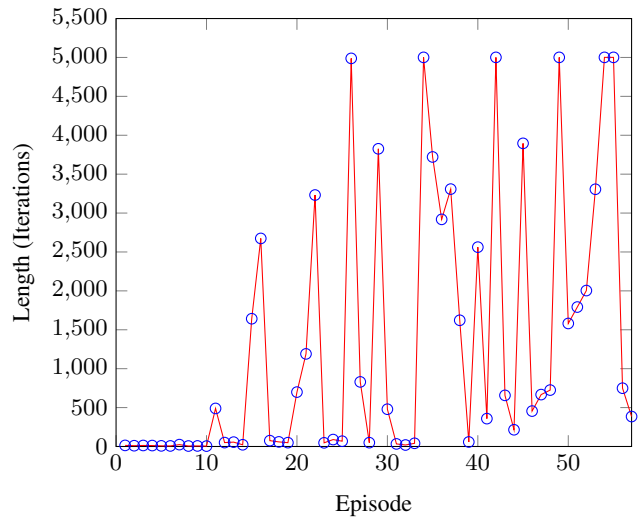

The full evolution of the episodes length during a single training can be seen on Fig.\ref{fig:hw_training}. The policy performance was low until enough samples were gathered; after that, the performance improved noticeably. 

During experimentation, the maximum episode length was limited to 5000 iterations ($\approx$ 7h). In all cases, this limit was reached after 30 episodes. In the training shown in Fig.\ref{fig:hw_training}, the threshold was reached on the 34th episode of the training.

The agent presents a high oscillation in the active CPUs to achieve temperature control. Although the learned policy is compatible with the main objective of maintaining the temperature below a threshold while maximizing the processing resources, this high oscillation is not desired as it produces unnecessary overhead when migrating the CPUs content (like processes and interrupts). A solution could be to penalize the CPU state changes in the reward calculation, which should force the agent to try to reduce this variability in the number of available CPUs if it is not needed.

Both the policy learning and the prediction with it was done without hardware acceleration (CPU only). During the policy learning, the amount of RAM used was stable in 290MB and spikes of 200 millicores (0.2 cores) were required every 5 seconds (prediction frequency).

\subsection{Limitations}
While with the current state definition the model is able to converge, initially, equivalent models with larger input dimensions were designed, and the model was unable to converge within a reasonable number of iterations. Although the usage of reinforcement learning for on-board optimization of satellite resources is promising, it could pose problems if the observation or action spaces are too large.

Starting from a model pre-trained in a synthetic environment with thermal simulation, 30 episodes were required to perform domain adaptation, during which the behavior of the agent was suboptimal. It is likely that when repeating tests in a real environment (i.e. outer space) a new domain adaptation will be necessary, and the agent’s behavior may be deficient one more time. Therefore, it is essential to accompany the system with a mechanism capable of protecting sensitive equipment, such as optical instruments, from inadequate policies during this adaptation period. For instance, the mechanism could involve shutting down the instrument or cooling the system when the temperature exceeds the acceptable limits.

\section{Conclusion}
In this work, a novel temperature management system called APaTheCSys was assessed on real-world space equipment that was operating on the ground. APaTheCSys was able to sustain a simulated heavy load for 12 hours without surpassing an enforced temperature threshold by constantly modifying the available processor resources.

Power control in this study was limited to the CPUs' processing power, but it might also encompass additional functions like switching off other devices or altering their energy modes. 

Overall, the method could be suitable for small payloads from satellites, spacecraft, or even planetary probes on upcoming space exploration missions, provided they are capable of hosting onboard intelligence and have the ability to control hardware from a software interface. This aligns with other research on autonomous systems for collision avoidance, docking, and debris removal and constitutes a step toward fully autonomous robotic missions.

\small
\bibliographystyle{plainnat}
\bibliography{bibFile}

\end{document}